\documentclass[letterpaper, 10 pt, conference]{ieeeconf}

\IEEEoverridecommandlockouts
\usepackage{cite}
\usepackage{amsmath,amssymb,amsfonts}
\usepackage{algorithmic}
\usepackage{graphicx}
\usepackage{textcomp}
\usepackage{xcolor}
\usepackage{siunitx}
\def\BibTeX{{\rm B\kern-.05em{\sc i\kern-.025em b}\kern-.08em
    T\kern-.1667em\lower.7ex\hbox{E}\kern-.125emX}}

\usepackage{xspace}
\usepackage{hyperref}
\usepackage{url}
\hypersetup{
    hidelinks,
    urlcolor=magenta,
}
\usepackage{cleveref}
\Crefname{equation}{Eq.}{Eqs.} 
\Crefname{section}{Sec.}{Secs.}
\Crefname{figure}{Fig.}{Fig.}
\Crefname{table}{Tab.}{tabs.}

\let\labelindent\relax      
\usepackage{enumitem}

\usepackage{siunitx}

\usepackage{balance}

\usepackage{tcolorbox}

\newcommand{\ra}[1]{\renewcommand{\arraystretch}{#1}}

\begin{document}

\title{Collision Probability Distribution Estimation \\via Temporal Difference Learning

\thanks{This research paper [project MORE] is funded by dtec.bw -- Digitalization and Technology Research Center of the Bundeswehr. dtec.bw is funded by the European Union -- NextGenerationEU. \\
All authors are with the Institute for Autonomous Systems Technology (TAS), Department of Aerospace Engineering, University of the Bundeswehr Munich, Neubiberg, Germany. Contact author email:
        {\tt\small
        thomas.steinecker@unibw.de
        }
}
}

\author{Thomas Steinecker, Thorsten Luettel and  Mirko Maehlisch}

\maketitle


\newcommand{\carla}{\textit{CARLA}\xspace}
\newcommand{\myname}{\textsc{CollisionPro}\xspace}
\newcommand{\timehorizon}{T\xspace}


\begin{abstract}
We introduce \myname, a pioneering framework designed to estimate cumulative collision probability distributions using temporal difference learning, specifically tailored to applications in robotics, with a particular emphasis on autonomous driving. This approach addresses the demand for explainable artificial intelligence (XAI) and seeks to overcome limitations imposed by model-based approaches and conservative constraints. We formulate our framework within the context of reinforcement learning to pave the way for safety-aware agents. Nevertheless, we assert that our approach could prove beneficial in various contexts, including a safety alert system or analytical purposes. A comprehensive examination of our framework is conducted using a realistic autonomous driving simulator, illustrating its high sample efficiency and reliable prediction capabilities for previously unseen collision events. The source code is publicly available.
\end{abstract}

\small
\textbf{\textit{Index Terms}---collision probability distribution, temporal difference learning, deep learning, robotics, autonomous driving, risk assessment, safety, safety awareness, explainable AI, reinforcement learning}

\urlstyle{sf}
\begin{center}

    \url{https://github.com/UniBwTAS/CollisionPro}
\end{center}

\section{Introduction} 

\begin{figure}[h!]
    \centering    
    \includegraphics[width=0.94\columnwidth]{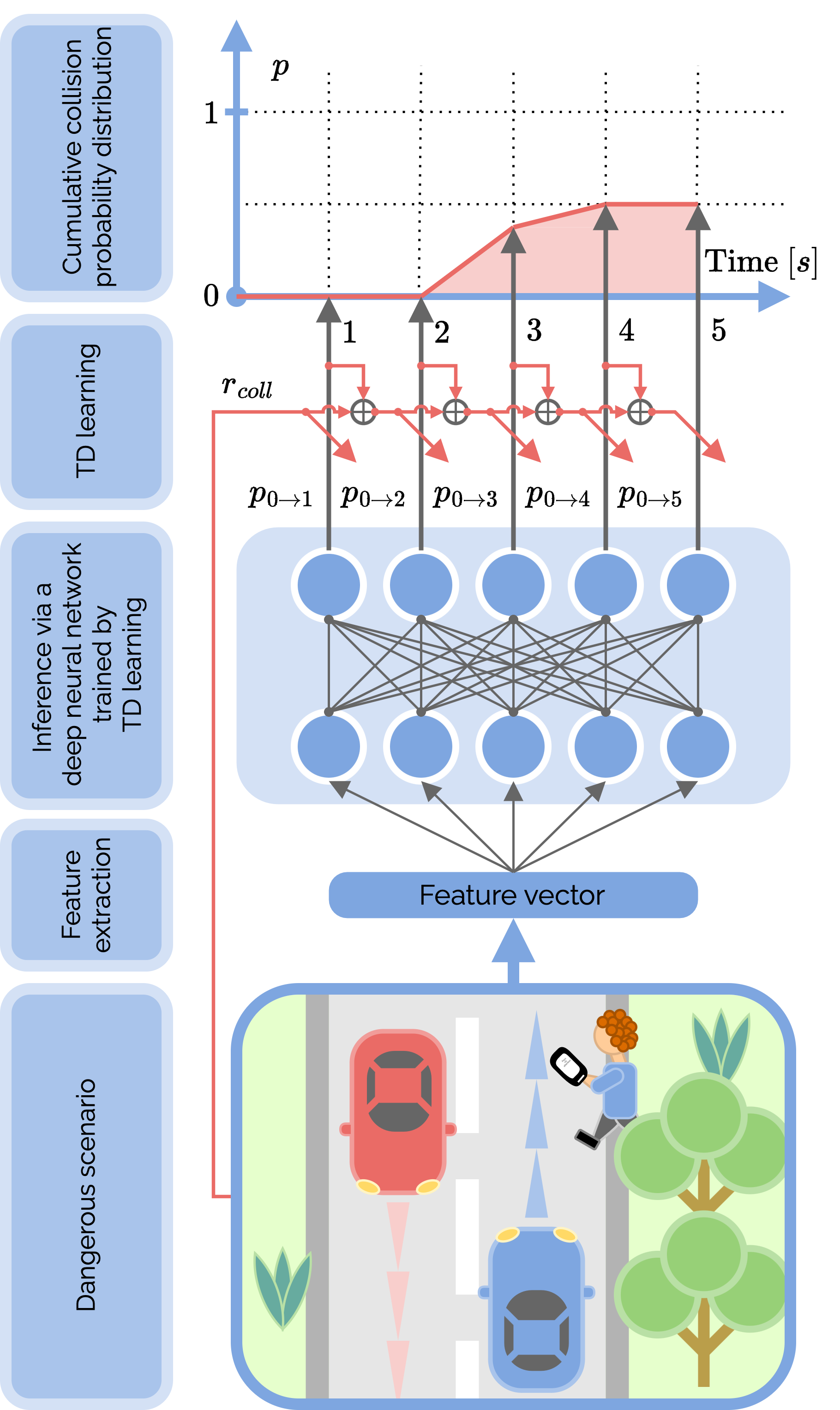}
    \caption{Visualization of the concept of \myname. Given a scenario, a feature vector is extracted, that is passed to a deep neural network (DNN), that was trained via temporal difference (TD) learning. The output is a vector of cumulative collision probabilities up to the specified time horizon, e.g. $T_H = \SI{5}{s}$. As can be seen in the figure, this sophisticated risk assessment strategy provides stochastic information about the true collision probability ($\sim0.5$) and the time where the collision is most likely ($\SIrange{2}{4}{s}$).
    Furthermore, the learning process is shown (red arrows), which is based on the principle of temporal difference learning. While the first estimator ($p_{0 \rightarrow 1}$) learns purely from collision and non-collision events/signals ($r_{coll}$), subsequent estimators ($p_{0 \rightarrow 2}$ to $p_{0 \rightarrow 5}$) can learn from all previous estimators (bootstrapping). A fundamental aspect of our approach lies in treating probabilities as equivalent to the value function.}
    \label{fig:overview}
    \vspace{-0.4cm}
\end{figure}


Fully understanding and preventing risk for autonomous driving is the key to its realization. Therefore we are convinced that a sophisticated risk assessment strategy is crucial for collision avoidance and consequently transport humans safely to their destinations.  


In this paper, we present \myname, a novel risk assessment framework that learns the cumulative collision probability distribution up to a specified time horizon via temporal difference learning. This is fundamentally different from most previous approaches that typically estimate a single collision probability value \cite{strickland2018deep, malawade2022spatiotemporal, althoff2009model, kibalov2020safe, althoff2011comparison} or qualitative risk values \cite{mokhtari2022don}. Estimating an entire distribution w.r.t. time comes with multiple benefits: Firstly, the temporal distribution of the collision probability provides a measure of the urgency and intensity of a preventive action. For example, if the collision probability is \SI{10}{\percent} after $3$ seconds, the system has more time to react compared to a potential collision after $1$ second, enabling a better trade-off between safety, performance, and comfort. Secondly, we can make use of temporal difference learning \cite{sutton2018reinforcement} by considering collision probabilities as the outputs of the value function in terms of reinforcement learning (RL). \Cref{fig:overview} illustrates an overview of our method. 


Our approach can be used in many ways. For example, to analyze a given strategy to better understand causes of collisions, but also to classify and analyze safe and unsafe scenarios. Moreover, the approach could be used as an alerting supervisor. It informs the autonomous driving system or human driver in case of an imminent collision, and consequently tries to reduce the collision probability proportional to the urgency provided by the distribution. Finally, it could also be interpreted as a value function and integrated into the training of an RL agent. In contrast to other agents, this would induce safety awareness in a physically interpretable manner. 

Safety is at the core of an autonomous system and is therefore represented by a wide range of approaches in both academia and industry. For instance, there are the rule-based approaches from MobileEye with responsibility-sensitive safety (RSS) \cite{shalev2017rss} and Nvidia with safety force field \cite{nister2019sff}. Both approaches are easy to apply and verifiable, but have their limits for uncertainties in perception and complex environments where the abstractions might fail to capture relevant characteristics.
 
The aforementioned approaches are concerned with avoiding collisions in advance by means of sophisticated rules and behavioral assumptions but are unaware/rudimentary aware of the scale of threat in a situation, whereas other approaches attempt to assess the danger of a situation in order to deduce appropriate actions. One promising and widely adopted approach is to predict the probability of a collision.

While our approach is rooted in machine learning, prior methods that also incorporated collision probability distributions \cite{philipp2019analytic, annell2016probabilistic} for risk assessment primarily relied on traditional techniques (refer to \Cref{tab:paper_classification}). In \cite{annell2016probabilistic} for example they calculate the Gaussian distributions of a collection of trajectories from another traffic participant and the trajectory of the ego vehicle by using Monte Carlo (MC) trials. In \cite{philipp2019analytic} they enhanced the performance over MC trials by approximating the rectangular shape with an octagon that allows for analytical calculations with negligible error.

In contrast to our approach, the concepts presented in \cite{philipp2019analytic, annell2016probabilistic} rely on substantial simplifications. These include the reduction of dynamic entities to simple geometries, and the behavior models of both the ego vehicle and other participants must be executed many times faster than the actual cycle time, making it challenging for complex, real systems. Another disadvantage of sample-based approaches is that for very small collision probabilities, they need a very large number of trials to make an accurate statement. Finally, the computational cost of these approaches depends highly on the number of dynamic entities.

Machine learning-based methodologies offer a more seamless approach to addressing the aforementioned limitations. Thus far, however, these approaches have primarily yielded single collision-oriented metrics. For instance, in \cite{strickland2018deep}, collision probability is determined with respect to a scenario, with an emphasis on ensuring the credibility of the value through approximated variational inference \cite{gal2016uncertainty}. Another notable method, as presented in \cite{malawade2022spatiotemporal}, involves determining a single collision value utilizing a scene graph that represents topological relationships. Nevertheless, no framework for collision probability distributions via learning-based approaches has been established to date. 
Our contributions can be summarized as follows:

\begin{itemize}
    \item[(1)] Introducing a generic framework leveraging temporal difference learning within the domain of reinforcement learning to estimate probability distributions of \textit{key} events, e.g. collisions.
    \item[(2)] Proposing \myname, an innovative deep learning architecture tailored to the estimation of collision probability distributions, aimed at enhancing both explainability in artificial intelligence and sample efficiency. 
\end{itemize}

The following part of the paper is structured as follows: First, in \Cref{sec:methods}, our generic framework for cumulative probability distribution is presented. In \Cref{sec:evaluation} the method is practically evaluated via an autonomous driving simulation environment and the results are discussed in \Cref{sec:discussion}. In \Cref{sec:conclusion} the paper is concluded. 

\begin{table}[!t]
\caption{Classification of collision probability approaches}
\centering
\ra{1.4}
\begin{tabular}{ c c c }
\hline
   & Single probability value & Probability distribution \\ 
 \hline
 Classical & \cite{althoff2009model, althoff2011comparison, kibalov2020safe} & \cite{philipp2019analytic, annell2016probabilistic} \\ 
 Learning-based & \cite{strickland2018deep, malawade2022spatiotemporal, candela2023risk} & ours \\ 
 \hline
\end{tabular}
\label{tab:paper_classification}
\vspace{-0.5cm}
\end{table}

\section{Methods}
\label{sec:methods}

In \Cref{sec:ccpde} the general framework for learning the cumulative collision probability is derived. Next in \Cref{sec:loss}, an appropriate loss function for a neural network approach is proposed. In \Cref{sec:performance} a performance measure is introduced that evaluates if the bootstrap technique is learning meaningful probability values. Finally, in \Cref{sec:rare_collisions} measures for handling rare collision events are discussed. 
In the following, the terminology for reinforcement learning is inspired by \cite{sutton2018reinforcement}.

\subsection{Cumulative Collision Probability Distribution Estimation}
\label{sec:ccpde}

In this section a novel approach for collision probability estimation is derived using temporal difference (TD) learning in the context of reinforcement learning (RL). 

We begin by defining the collision probability $p_{t \rightarrow t + N_H}$ for a specific time horizon $T_H = N_H \Delta T$, where $\Delta T$ represents the time step between two consecutive points in time, as follows:
\begin{equation}
p_{t \rightarrow t + N_H} = \frac{N_{\text{collisions within }T_H}}{N_{\text{collisions within }T_H} + N_{\text{non-collisions within }T_H}}.
\end{equation}
Here, $N$ denotes a counter of events. Note that the notation $t + i$ is a shortcut for $t + i \Delta T$. In order to reduce the complexity of the above equation we calculate the probability incrementally as 
\begin{equation}
    p_{t \rightarrow t + N_H} \xleftarrow{} p_{t \rightarrow t + N_H} + \alpha (c_{t \rightarrow t + N_H} - p_{t \rightarrow t + N_H}),
\end{equation}
where $\alpha$ is the step size for the incremental learning update, and $c_{t \rightarrow t + N_H}$ specifies whether a collision event occurred within the time interval $[t, t + T_H]$.

In order to preserve RL terminology we want to consider $c_{t \rightarrow t + N_H}$ as the \textit{return} which is typically defined as follows
\begin{equation}
    G_t = r_{t + 1} + \gamma r_{t + 2} + \ldots + \gamma^{T - 1} r_{T},
\end{equation}
where $r$ specifies the reward, $\gamma$ the discount factor and $T$ is the time at the end of the episode. 
In order to use the same terminology we suggest that a collision event refers to a reward of $1$ and $0$ otherwise and that the return is truncated after the time horizon $T_H$, which results in the following \textit{fixed-finite return}
\begin{equation}
    \mathcal{G}_{t \rightarrow t + N_H}^{\text{MC}} = r_{t+1} + r_{t+2} + \ldots + r_{t + N_H}.
    \label{eq:MC_return}
\end{equation}
Note, in case of a collision prior to the specified time horizon the subsequent rewards are not \textit{experienced} and are consequently $0$.

The equation above corresponds to a modified version of Monte Carlo (MC) learning, which results in high variance but unbiased updates. It is well known that temporal difference (TD) learning typically has better learning performance as it comprises bias and variance. We therefore strive to extend this approach to make use of the bootstrapping mechanism. However, contrary to an original RL approach we cannot bootstrap the probability estimation $p_{t \rightarrow t + N_H}$ of future states as they incorporate information beyond the time horizon of the current state. To resolve that issue we introduce multiple probability estimators  equidistantly in-between the interval $[0, T_H]$
\begin{equation}
    p_{t \rightarrow t + i} \:\: \mid \:\: i \in {1, ..., N_H}.
    \label{eq:multiple_predictors}
\end{equation}

Henceforth, we will use the terms estimator and head interchangeably.

\begin{figure}[t!]
    \centering    
    \includegraphics[width=\columnwidth]{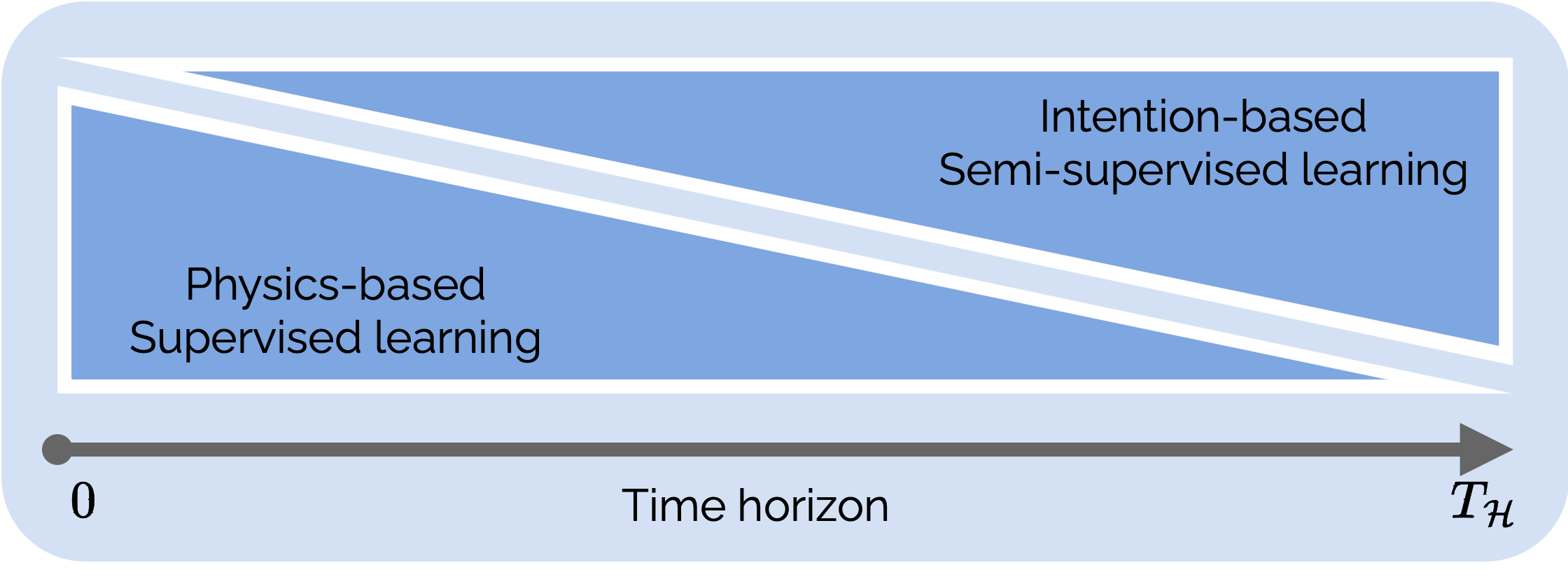}
    \caption{The relationship between long-term and short-term predictions (similar to \cite{lefevre2014survey}). Whereas for short-term predictions the current kinematics and dynamics is crucial for risk assessment, intentions of all dynamic agents become increasingly important for long-term predictions.}
    \label{fig:horizon}
\end{figure}

Next, we generalize the return from \Cref{eq:MC_return} for all estimators as defined in \Cref{eq:multiple_predictors} to the $n$-step return 
\begin{equation}
        \mathcal{G}_{t \rightarrow t + i}^{n} = r_{t + 1} + r_{t + 2} + \ldots + r_{t + n - 1} + p_{t + n \rightarrow t + i},
        \label{eq:TD_return}
\end{equation}
where $n$ is a value between $1$ and $i$ and for $i = n$ we obtain the Monte Carlo return as described in \Cref{eq:MC_return}. Note that for $\mathcal{G}_{t \rightarrow t + 1}$, bootstrapping is not meaningful, as the agent is in a collision in the next time step or not. Consequently, we use the direct reward signal $r_{t+1}$. This is similar to the final state handling in episodic RL approaches. 

By introducing new probability estimations we accomplished two beneficial characteristics: We obtain a collision probability distribution up to the specified time horizon and we can apply the bootstrapping mechanism which was defined in the equation above.

Now that we formulated the problem in the context of RL we can further improve the learning performance by using TD$(\lambda)$, meaning that we do not bootstrap from a single $n$-step return as described in \Cref{eq:TD_return}, but from all of them by introducing the $\lambda$-return
\begin{equation}
        \mathcal{G}^{\lambda}_{t \rightarrow t + i} = \sum_{n=1}^{i} \lambda_n \mathcal{G}_{t \rightarrow t + i}^n,
    \label{eq:lambda_return}
\end{equation}
where $\lambda$ is defined as follows 
\begin{equation}
    \lambda_n = 
    \begin{cases}
      (1 - \lambda) \lambda^{n} & \text{if $n < i$}\\
      1 - (1 - \lambda) \sum_{j=1}^{t + i - 1} \lambda^{n - 1} & \text{else}.
    \end{cases}       
\end{equation}

Finally, we can make use of the $\lambda$-return described in \Cref{eq:lambda_return} to formalize the update equation for all estimators as described in \Cref{eq:multiple_predictors}: 
\begin{equation}
    p_{t \rightarrow t + i} \xleftarrow{} p_{t \rightarrow t + i} + \alpha (\mathcal{G}^{\lambda}_{t \rightarrow t + i} - p_{t \rightarrow t + i}).
\end{equation}

\begin{figure*}[ht!]
    \centering    
    \includegraphics[width=2\columnwidth]{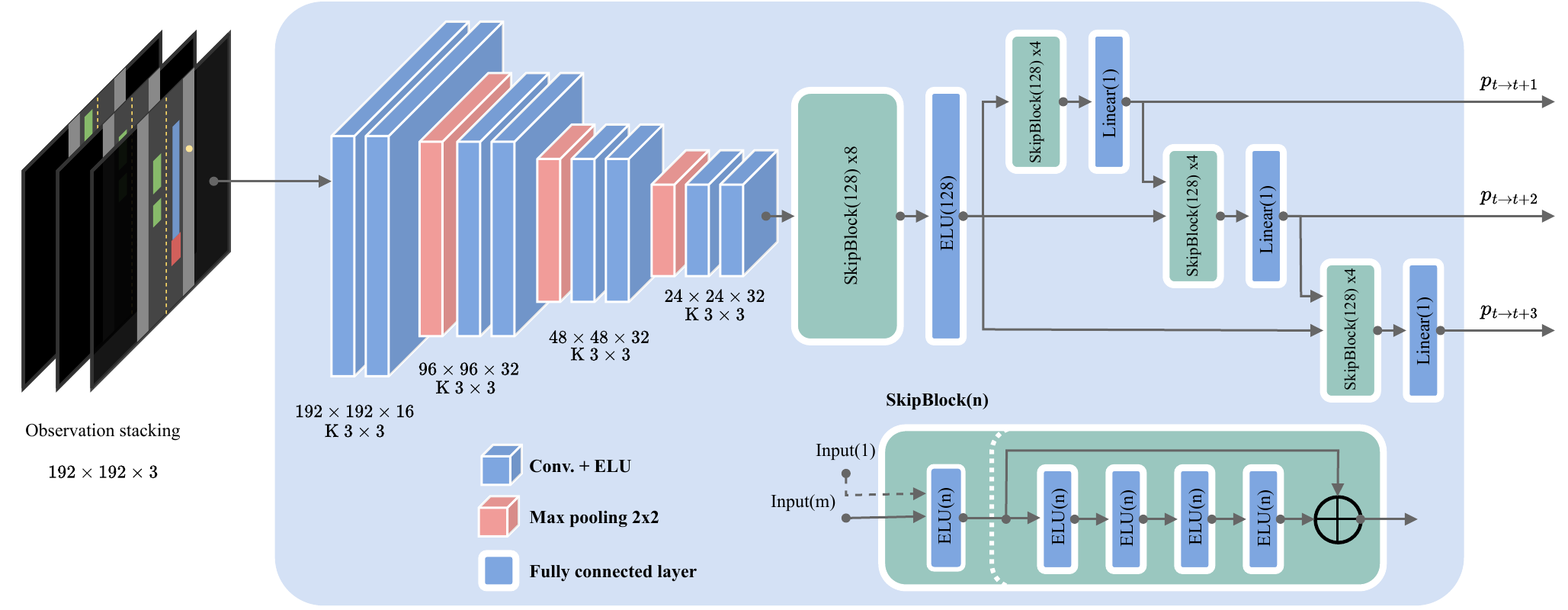}
    \caption{The network architecture and pipeline for learning the cumulative collision probability distribution. \textbf{Input}: The input consists of a stacked observation space of the bird's-eye semantic view transformed to greyscale, resulting in an input dimension of $192 \times 192 \times 3$ considering three consecutive time steps. Due to hardware limitations, three time steps were chosen as this allows the dynamics of all entities to be captured, but further time steps into the past would presumably produce better predictions, as these could provide a richer representation of behavioral patterns and intentions. \textbf{Architecture}: The network consists of three parts: The encoder which consists of a sequence of convolutional neural network (CNN) layers (K $3 \times 3$ indicates the kernel matrix size) that are further processed by a fully connected block. The two aforementioned network components form the common backbone for the individual sub-networks that use the latent space from the backbone and the output from its predecessor head. The network is illustrated for $H=3$ heads/estimators. The initial \textit{SkipBlock} within a sequence of \textit{SkipBlocks} is denoted by the solid line in the bottom-right, tasked with transforming the inputs to the desired dimension. Subsequent blocks are represented by the dashed line.}
    \label{fig:architecture}
\end{figure*}

In \Cref{eq:multiple_predictors}, we introduced multiple heads to leverage the bootstrapping mechanism. We aim to further justify this strategy with \Cref{fig:horizon}. A smaller horizon is primarily influenced by physical properties, while a larger horizon is predominantly influenced by the intentions of dynamic entities. However, in reasoning about the outcome of intentions, physics plays a crucial role shortly before a collision.

Finally, we would like to point out that we set the collision reward to $-1$ instead of $1$ to be consistent with the RL approach, which was formulated as a maximization optimization. This has no further impact on the framework derived above, except that the signs of the probability estimators have to be negated to obtain plausible probability values in the interval $[0, 1]$. 

\subsection{Loss}
\label{sec:loss}

The \textit{cumulative} collision probabilities are approximated by a neural network (NN). The outputs of the NN should be compliant with the following constraints of a meaningful cumulative probability distribution:
\begin{itemize}[leftmargin=1.3cm, topsep=8pt, itemsep=3pt, labelsep=0.35cm]
    \item[($P_1$)] $p_{t \rightarrow t + i} \in [0, 1]$ 
    \item[($P_2$)] $p_{t \rightarrow t + i} \leq p_{t \rightarrow t + j} \:\: \forall{i, j} \:\: i < j$ \: .
\end{itemize}

Next, we want to discuss the loss function, which consists of three parts (we indicate predicted values with a hat symbol):

\begin{itemize}[leftmargin=1.3cm, topsep=8pt, itemsep=3pt, labelsep=0.35cm]
    \item[($L_1$)] The error with respect to the target value which is the return value defined in \Cref{eq:lambda_return}. For the choice of a meaningful loss, it is necessary to ensure that it is consistent with definition of probability. The popular \textit{mean squared error} (MSE) is a suitable candidate for this, since it finds a minimum for the expected value of a random variable. It can be formulated for one sample as follows:
    \begin{equation}
        \L_{\text{MSE}} = (\mathcal{G}^{\lambda}_{t \rightarrow t + i} - \hat{p}_{t \rightarrow t + i})^2.
        \label{eq:L1}
    \end{equation}
    \item[($L_2$)] The error with respect to $P_1$. Deviations should be punished for estimates that fall outside  the valid \textit{interval}, and we achieve this by applying a linear penalty proportional to the minimum distance from the interval $[0, 1]$ ($c_I$ is a weighing factor):
    \begin{equation}
        \L_I = c_I \: \text{max}(0, \text{max}(-\hat{p}_{t \rightarrow t + i}, \hat{p}_{t \rightarrow t + i}-1)).
    \end{equation}
    \item[($L_3$)] The error with respect to $P_2$. Ultimately, $L_3$ penalizes an estimate that is smaller than the predecessor and/or larger than the successor, which aligns with the definition of cumulative distributions ($c_C$ is a weighing factor): 
    \begin{equation}
    \hspace*{-0.5cm}
    \L_C = c_C
    \begin{cases}
      \text{max}(\hat{p}_{t \rightarrow t + i} - \hat{p}_{t \rightarrow t + i + 1}, 0) & \text{if $i = 1$}\\
      \text{max}(\hat{p}_{t \rightarrow t + i - 1} - \hat{p}_{t \rightarrow t + i}, 0) & \text{if $i = N_H$}\\
      \text{max}(\hat{p}_{t \rightarrow t + i - 1} - \hat{p}_{t \rightarrow t + i}, 0) \\+ \text{max}(\hat{p}_{t \rightarrow t + i} - \hat{p}_{t \rightarrow t + i + 1}, 0) & \text{else.}
    \end{cases}       
    \end{equation}
    \label{eq:L3}
\end{itemize}

\subsection{Performance Measure}
\label{sec:performance}

The loss function defined in \Cref{sec:loss} may not be utilized as a performance measurement for accuracy, since the error calculation is based on bootstrapping, so that the error refers to estimated values, which are biased.

Instead, the return as defined in \Cref{eq:MC_return} can be used as a reference, having high variance, but no bias. For the error calculation the MSE is suitable, since it's minimum coincides with the expected value, providing consistency with the definition of probabilities. The measure of performance can then be given as follows:
\begin{equation}
    \mathcal{E}_{\text{acc}} = (\mathcal{G}_{t \rightarrow t + i}^{\text{MC}} - \hat{p}_{t \rightarrow t + i})^2.
\end{equation}

Another relevant characteristic is over- or underestimation of the total collision probability estimation. An overestimate corresponds to \textit{pessimism}, since this generally provides collision estimates that are higher than the actual probability. For instance, if an neural network would fail to learn meaningful estimates and provides a collision probability of $0$ for each state, this would be very optimistic for a realistic environment. Pessimism can be assessed by taking the signed distance between the unbiased return and the estimate:
\begin{equation}
    \mathcal{E}_{\text{pes}} = \mathcal{G}_{t \rightarrow t + i}^{\text{MC}} - \hat{p}_{t \rightarrow t + i}.
\end{equation}

\subsection{Handling Rare Collision Events}
\label{sec:rare_collisions}

In realistic environments, collisions are rare, even when including extremely challenging scenarios. However, to ensure efficient learning, a more balanced distribution between collision and non-collision data is essential. Often, prioritization methods like Prioritized Experience Replay (PER) \cite{schaul2015prioritized} are employed in the RL domain, typically using the TD-error for prioritization assessment. In our approach, we incorporate the concept of PER alongside undersampling, as discussed in \cite{devi2020review}, to tackle the imbalance between collision and non-collision data. This involves sampling collision-related samples with probability $p_c$ and non-collision-related samples with probability $p_{nc}$. Analogous to PER, this sampling strategy requires proportional weighting, necessitating modifications to the loss function, as defined in \Cref{eq:L1}: 
\begin{equation}
    \L_{\text{MSE}} = 
    \begin{cases}
      (\mathcal{G}^{\lambda}_{t \rightarrow t + i} - \hat{p}_{t \rightarrow t + i})^2 & \text{\textit{non-collision}}\\
      \frac{p_{nc}}{p_{c}} (\mathcal{G}^{\lambda}_{t \rightarrow t + i} - \hat{p}_{t \rightarrow t + i})^2 & \text{\textit{collision.}}
    \end{cases}       
\end{equation}

\section{Evaluation}
\label{sec:evaluation}

\subsection{Settings}
\label{sec:settings}

In the following, we will describe the simulation environment, the input of the neural network, the architecture of our approach, and the training setup.

\textbf{Environment}---To evaluate \myname, we employed the \carla simulator \cite{carla}, along with the traffic manager (TM). In each episode, we introduced $30$ vehicles, $30$ pedestrians, and the ego vehicle into \textit{Town02}. The TM controlled all dynamic entities, excluding the ego vehicle. To enhance the challenge, vehicle autopilots frequently exhibited non-compliance with traffic rules, such as excessive speeding, disregarding traffic lights, or failing to maintain a safe distance from preceding vehicles. Pedestrian speeds were uniformly distributed between \SIrange{1}{5}{\meter\per\second}.

The ego vehicle adhered to a straightforward control strategy, complying with traffic lights while exceeding the maximum velocity by \SI{25}{\percent}. Additionally, the offset from the center of the lane was uniformly chosen within a deviation of \SI{25}{\centi\meter} on both sides. Collision checking for the ego vehicle involved predicting other vehicles and itself \SI{1.5}{\second} into the future, using the desired ego path and a linear prediction of other vehicles. In the presence of a potential collision or upcoming curves, the ego vehicle applied brakes.

The dynamic environment's versatile behavior settings aimed to create more realistic scenarios, where the actions of other entities were not entirely observable or predictable, necessitating a learning approach based on sufficient experience.

\textbf{Input and Observation Space}---The observation space comprises a semantic bird's-eye view based on \cite{chen2019deep}. It extracts high-level information from the road topology (lanes and markings) presented in the OpenDRIVE format within \carla. The ego vehicle and all other vehicles are represented as rectangular approximations, while walkers are approximated by circles, along with the inclusion of the ego vehicle's desired path. To simplify the state representation complexity, we apply a grayscale transformation. Observations are stacked over multiple time steps to incorporate the dynamics of the environment. A representative example of the observation space is depicted in \Cref{fig:architecture} on the left side.

\textbf{Network Architecture}---The neural network is depicted in \Cref{fig:architecture}. The input comprises a stacked semantic bird's-eye view, as discussed above. Through successive CNN layers, the images are encoded, and key features are extracted. Subsequently, the output of the last CNN layer is directed to the second part of the common backbone, which comprises fully connected layers and skip connections.
The output of the backbone is then passed to each sub-network assigned to estimate a specific cumulative collision probability. Furthermore, each sub-network of the head $p_{t \rightarrow t + i}$ receives the output of its predecessor $p_{t \rightarrow t + i - 1}$, reinforcing the constraint that cumulative probabilities strictly increase, as also penalized by the loss function described in \Cref{sec:loss}. 

\begin{figure*}[ht!]
    \centering    
    \includegraphics[width=2\columnwidth]{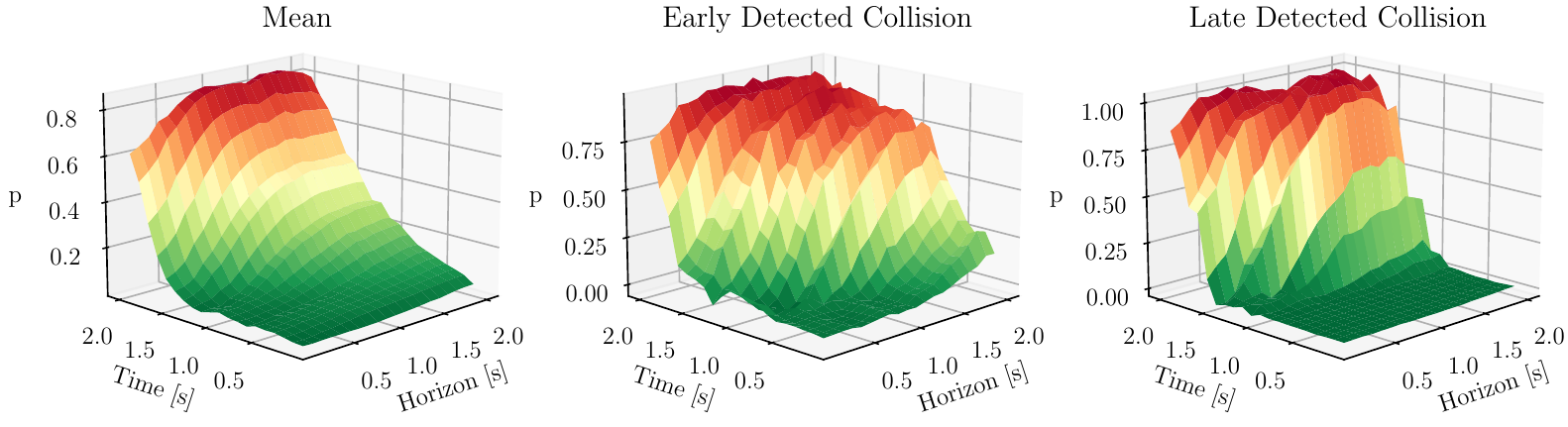}
    \caption{Collision characteristics: The collision probability distribution in the last $T_{H}=\SI{2.0}{\second}$ before the event of a collision. \textbf{Left}: The mean collision characteristic over 50 collision scenarios. \textbf{Center}: An example scenario in which the collision was predicted at an early stage with a high degree of confidence. \textbf{Right}: An example scenario in which the collision was only predicted with confidence a short time beforehand (approximately $1$ second).}
    \label{fig:collision_characteristics}
\end{figure*}

\begin{figure}[t!]
    \centering    
    \includegraphics[width=\columnwidth]{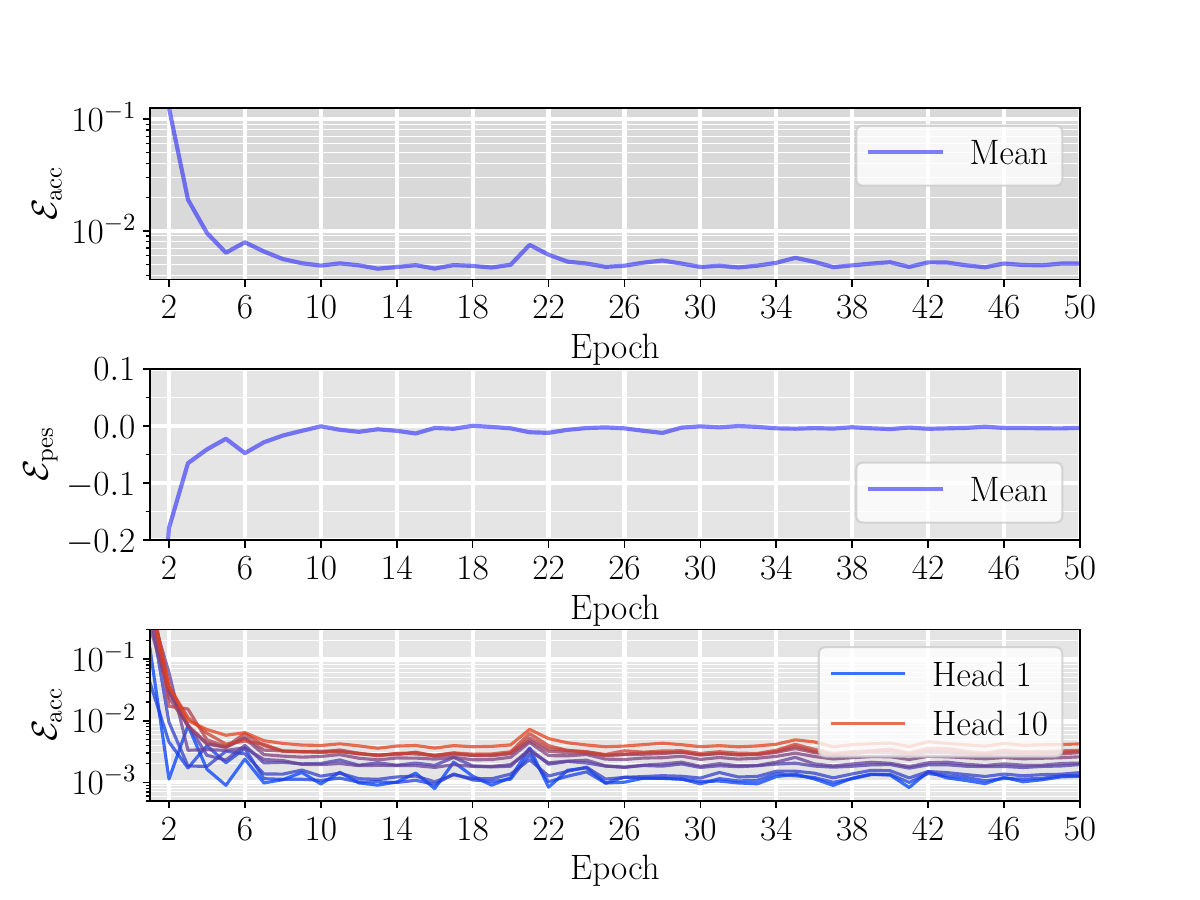}
    \caption{Error plots over $50$ epochs with $6 \cdot 10^3$ samples. \textbf{Top}: Mean error $\mathcal{E}_{\text{acc}}$ over all heads (see performance measure in \Cref{sec:performance}). \textbf{Center}: Mean error $\mathcal{E}_{\text{pes}}$ over all heads of the performance measure. \textbf{Bottom}: Shows the accuracy performance $\mathcal{E}_{\text{acc}}$ for each individual head, where head $i$ corresponds to $p_{t \rightarrow t + i}$.}
    \label{fig:error_plots}
\end{figure}

\begin{figure}[t!]
    \centering    
    \includegraphics[width=0.9\columnwidth]{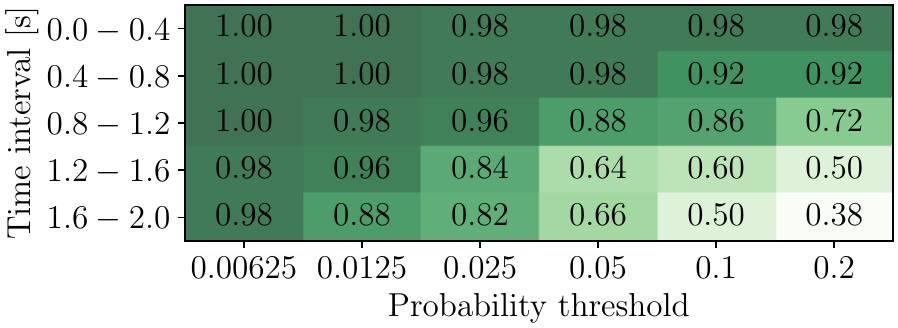}
    \caption{Detection rate of collisions within a time interval preceding the collision event and probability threshold for the final head ($p_{t \rightarrow t + N_{H}}$) across 50 collision events. We emphasize that converting a probability distribution into a binary value (collision or non-collision) for the implementation of our framework undermines the added value of our approach towards collision detection/handling. The plot presented above serves merely as a simplification for visualizing the results.}
    \label{fig:collision_table}
\end{figure}

\textbf{Training Setup}---The learning pipeline comprises two phases: Episode generation and learning via temporal difference. Initially, we generated $2000$ episodes with $\Delta T = \SI{0.1}{\second}$ and a maximum episode length of $3000$ steps. Episodes concluded either due to reaching the maximum steps, a collision occurrence, or a blocking scenario (where the ego vehicle remains stationary for more than \SI{45}{\second}), resulting in $744$ collision scenarios.

After generating all episodes, samples were randomly drawn with $p_c=0.25$ and $p_{nc}=0.025$. For each epoch, 6000 samples were evaluated with a batch size of $2$ over a total of $50$ epochs. Optimization employed the Adam optimizer \cite{kingma2014adam} with exponential learning decay from $10^{-5}$ to $10^{-6}$. The weighting for the loss, as defined in \Cref{sec:loss}, is $c_I = 1$ and $c_C = 1$.

We specified the time horizon as $T_{H}=\SI{2.0}{s}$, resulting in $20$ heads ($N_H = T_{H} / \Delta T = \SI{2.0}{s} / \SI{0.1}{s} =  20$). In \Cref{fig:architecture}, only $3$ heads are illustrated for clarity. The $\lambda$-return, as defined in \Cref{eq:lambda_return}, is calculated using $\lambda=0.8$ and truncation after $n=10$.

\subsection{Results}
\label{sec:results}

In the following, the results from the outcome of the training as described in \Cref{sec:settings} are described. In \Cref{fig:error_plots}
the performance measures as discussed in \Cref{sec:performance} are shown for unseen episodes. The subfigures at the top and at the middle show an effective learning progress. Especially the convergence of $\mathcal{E}_{\text{pes}}$ shows that the overall learning procedure is unbiased. At the bottom of \Cref{fig:error_plots}
the accuracy for the individual heads is shown. With increasing prediction horizon the performance decreases significantly. This is a consequence of the increasing complexity of the task of heads with longer prediction horizon, e.g. the task of the first head is to consider the next time step and events that happen later are independent of that estimation. The last head on the other hand has to consider the overall time horizon. In \Cref{fig:horizon}, this complexity across the prediction horizon was already implied.

\Cref{fig:collision_characteristics} depicts the estimated collision probability over the last $T_{H}$ seconds. The subfigure on the left displays the mean characteristic over $50$ unseen collisions. While predictions exhibit confidence shortly before the collision event, they tend to diminish further away from the event. This could be attributed to two possible reasons: either the network was less efficient in learning the true probability, or, due to the stochastic nature of the unobservable and multi-agent environment, the observation stack, as detailed in \Cref{sec:settings}, may not capture all the essential information. We contend that the latter factor plays a crucial role, influenced by the low resolution and unpredictable behavior of dynamic entities, making precise predictions unattainable. In the center of \Cref{fig:collision_characteristics}, an example is shown where enough information was provided by the state representation to confidently predict the collision. In contrast, the subfigure on the right demonstrates an example where enough evidence was only perceptible shortly before the collision event. These insights could be utilized for further analyzing the quality of the state representation and comparing different feature selection strategies.

Finally, we present insights in \Cref{fig:collision_table} regarding the collision detection rate at different thresholds for the probability of the last head ($p_{i \rightarrow i + N_H}$) and time intervals ahead of the collision. It is demonstrated that with small probability thresholds, such as $0.0125$, and shortly before the collision event (e.g., $0.0-\SI{0.4}{\second}$), the detection capability is quite high but significantly decreases with a longer prediction horizon and higher thresholds. Since our framework offers a probabilistic collision detection distributed over time, we argue that a link can be established between the necessity for action and the utilization of the probabilistic output. This allows for the consideration of the actual probability values, where higher values, especially when closer to the present time, indicate an increased urgency for action, and vice versa. We would like to note that the specification of false or true negative errors do not exist in our probabilistic framework, as the non-occurrence of collisions at a given non-zero probability does not imply a discrepancy with the estimated probabilities in the respective situation. Consequently, for a fair assessment of the algorithm's performance, please refer to \Cref{fig:error_plots}.

\section{Discussion}
\label{sec:discussion}

In \Cref{sec:evaluation}, we showcased the practicality of this approach using \carla, an autonomous driving simulator. The algorithm was supplied solely with stacked semantic bird's-eye views as observations and a collision detection signal. Remarkably, with fewer than $1000$ observed collisions, the algorithm demonstrated reliable predictions of unseen collision scenarios, underscoring its significant sample efficiency.

In recent years, there has been a growing emphasis on explainable artificial intelligence (XAI) due to the perceived drawback of its black-box nature \cite{confalonieri2021historical}. Our approach contributes to this effort in an elegant manner, as the outcome inherently provides an explanation. The probabilities can be viewed as values akin to reinforcement learning, where the probability is analogous to the negative of the value function. This implies that the value function itself is interpretable, resembling reinforcement learning (RL) for games like in \textit{AlphaZero} \cite{silver2017mastering}. By employing our formalism, we believe it is conceivable to develop safety-aware agents that possess the ability to articulate decisions through a deeper comprehension of safety.

In the evaluation outlined in \Cref{sec:evaluation}, we simulated aggressive driving behavior, leading to collision rates significantly exceeding those observed in real-world driving scenarios by several orders of magnitude. This raises questions about the scalability of our approach to increasingly rare collisions. In this regard, we would like to introduce three considerations: (1) We can mitigate rarity through our sampling strategy, as presented in \Cref{sec:rare_collisions}. (2) Training in challenging environments—characterized by factors such as poor perception, significant occlusions, high traffic volumes, and aggressive driving behaviors—may lead to improved robustness, as demonstrated in a similar analogy presented in \cite{ha2018world}. However, this may result in an overestimation of probabilities (pessimistic estimation). (3) By selectively choosing scenarios—such as complex intersections or adverse weather conditions—we can provoke collisions without directly influencing the true collision probability. This is because collision probabilities are estimated based on their environments and history, rather than the absolute collision frequency.

Thus far, our approach to value function approximation has been outlined within the context of a predetermined policy. However, this framework may seem limiting, as it implies an inevitability of collision under such a policy. However, the application or further development of our approach plays a crucial role. For instance, if implemented as a warning system for a human driver, it would prompt the driver to take suitable actions in response to warning signals, necessitating deviations from the policy employed during the learning phase. Integrating \myname into a RL agent would dynamically alter both the policy and collision estimation throughout the learning process, given their direct interdependence, akin to the behavior observed in a typical value function. Finally, in future work, we aim to supplement \myname with Q-values so that the collision probability distribution depends on the action. This modification would obviate the necessity of relying on RL methodologies like actor-critic \cite{konda1999actor} to leverage the value function, offering direct collision probabilities for individual actions. However, it's worth noting that this approach is restricted to discrete action spaces and may find greater utility in high-level action spaces.

\section{Conclusion}
\label{sec:conclusion}

We introduce \myname, a sample-efficient framework for estimating cumulative collision probability distributions. Utilizing a modified temporal difference approach, we leverage the bootstrapping mechanism, transforming the task into a semi-supervised approach, with feedback provided solely on the occurrence of a collision event. We evaluate \myname in an autonomous driving simulator, demonstrating reliable collision predictions for previously unseen events. Given its simplicity and generic nature, we believe that our approach could be applied to various applications, making a significant contribution to explainable AI.

\bibliographystyle{IEEEtran}  \balance 
\bibliography{macros/IEEEabrv,macros/additional_abrv,macros/et_al,ref}

\end{document}